\documentclass{article}

\usepackage[includeheadfoot,
            bindingoffset=0mm,
            inner   = 25mm,
            top     = 10mm,
            outer   = 25mm,
            bottom  = 10mm,
            paperwidth = 210mm,
            paperheight = 297mm,
            ]{geometry}
\usepackage[margin={2.0cm,0cm},oneside,labelfont={sf,bf},singlelinecheck=false]{caption}

\usepackage{graphicx}

\usepackage[fleqn]{amsmath}
\usepackage{accents}
\usepackage{amssymb}
\usepackage{amsmath}
\usepackage{tabularx}

\usepackage{enumitem}
\usepackage{cite}
\usepackage[perpage,multiple]{footmisc}
\usepackage{hyphenat}
\usepackage[english]{babel}
\usepackage[section]{placeins}
\usepackage{color}
\usepackage{upgreek}
\usepackage{listings}

\definecolor{codegreen}{rgb}{0,0.6,0}
\definecolor{codeblue}{rgb}{0,0,0.8}
\definecolor{codegrey}{rgb}{0.5,0.5,0.5}
\usepackage{amsthm}
\theoremstyle{definition}

\usepackage{titlesec}
\titleformat{\section}[hang]{\Large\bfseries\raggedright\sffamily}{\thesection}{1em}{}
\titleformat{\subsection}[hang]{\large\bfseries\raggedright\sffamily}{\thesubsection}{1em}{}
\titleformat{\subsubsection}[hang]{\normalsize\bfseries\raggedright\sffamily}{\thesubsubsection}{1em}{}
\usepackage{abstract}

\usepackage{authblk}

\begin{document}

\title{ \huge\bfseries\sffamily Probabilistic Kolmogorov-Arnold Network }

\author[1]{Andrew Polar}
\author[2,3,4]{Michael Poluektov}
\affil[1]{Independent software consultant, Duluth, GA, USA}
\affil[2]{Department of Mathematical Sciences and Computational Physics, School of Science and Engineering, University of Dundee, Dundee DD1 4HN, UK}
\affil[3]{IINM, WMG, University of Warwick, Coventry CV4 7AL, UK}
\affil[4]{Corresponding author, email: mpoluektov001@dundee.ac.uk}

\date{ \huge\normalfont\sffamily DRAFT: \today }

\maketitle

\setlength{\absleftindent}{2.0cm}
\setlength{\absrightindent}{2.0cm}
\setlength{\absparindent}{0em}
\begin{abstract}
The Kolmogorov-Arnold network (KAN) is a regression model that is based on a representation of an arbitrary continuous multivariate function by a composition of functions of a single variable. Experimentally-obtained datasets for regression models typically include uncertainties, which in some cases, cannot be neglected. The conventional way to account for the latter is to model confidence intervals of the systems' outputs in addition to the expected values of the outputs. However, such information may be insufficient, and in some cases, researchers aim to obtain probability distributions of the outputs. The present paper proposes a method for estimating probability distributions of the outputs in the case of aleatoric uncertainty (i.e. for systems that produce different outputs each time an experiment is executed with the same inputs). The suggested approach covers input-dependent probability distributions of the outputs and is capable of capturing the multi-modality, as well as the variation of the distribution type with the inputs. Although the method is applicable to any regression model, the present paper combines it with KANs, since the specific structure of KANs leads to computationally-efficient models' construction. The source code is available online.
\end{abstract}

\section{Introduction}
\label{sec:intro}

The popularity of using the Kolmogorov-Arnold representation \cite{Arnold1957,Kolmogorov1956} as a regression model has significantly grown over the last months, largely due to publication of preprint \cite{Liu2024}, which promotes term \emph{network} (leading to abbreviation `KAN'). However, the idea of using the Kolmogorov-Arnold representation as a machine-learning model has been out for decades \cite{Bryant2008,Liu2015}, and successful implementations of this model and its training method have been available for years \cite{Sprecher1996,Sprecher1997,Koeppen2002,Igelnik2003,Coppejans2004,Actor2017,Actor2018,Deventer2022}, including the developments by the authors of the present paper \cite{Polar2021,Poluektov2023}.

Most recent preprints focus on deterministic modelling, and only few authors tried using this model for estimating probabilistic properties. For example, in \cite{Duda2024}, a method for propagating statistical moments through the model is suggested, while in \cite{Hassan2024}, KANs are combined with Bayesian inference for estimating posterior distributions. The present paper proposes an ensemble training method for uncertainty quantification and combines it with KANs.

There is a large variety of ensemble training methods \cite{Lakshmi2017}, such as bagging \cite{Efron1979,Breiman1996}, boosting \cite{Schapire1990,Freund1997}, stacking \cite{Wolpert1992,Smyth1999}, random forest \cite{Breiman2001}, and multiple variations of these methods. Such methods create a set of models, which are used to calculate a set of outputs for one given input of a considered system. The set of outputs is then used for estimating the uncertainty. In the simplest scenario, that can be a confidence interval for the true system output. The methods above can be used in combination with any regression model. There are also methods tailored to specific models, such as Monte Carlo dropout \cite{Gal2016}, which is tailored to neural networks. 

In the present paper, a new ensemble training method is proposed --- the \emph{divisive data re-sorting} (DDR). As will be shown below, ensembles obtained using DDR can be converted to probability distributions of the system output. Such distributions may be of an arbitrary type (not necessarily bell-shaped) --- the method captures multi-modality of the distributions. The method is then combined with the Kolmogorov-Arnold model's architecture resulting in a \emph{shallow probabilistic model}.

\section{Divisive data re-sorting (DDR)}
\label{sec:DDR}

A dataset is considered to contain independent data records $\left( X^i, y_i \right)$, $i \in \left\lbrace 1, \ldots, N \right\rbrace$, where $X^i \in \mathbb{R}^m$ is the input of the $i$-th record, $y_i \in \mathbb{R}$ is the output of the $i$-th record, and $N$ is the number of records. The records are observations of a system with aleatoric uncertainty --- the output of such system is a random variable with input-dependent probability distribution. 

Suppose some deterministic regression model $M_0: \mathbb{R}^m \to \mathbb{R}$ can be built using a particular error minimisation process,
\begin{equation*}
  M_0 = \underset{M\in\mathcal{A}}{\operatorname{argmin}} \sum_i \left( y_i - M\left(X^i\right) \right)^2 ,
\end{equation*}
where $\mathcal{A}$ is the space of possible models. Such models will be called \emph{expectation models}. 

The method starts by building one expectation model for the entire dataset $M_{1,1}$, where the first subscript is the step number and the second subscript is the index of the model within the step. Then, the data records are re-sorted according to residuals
\begin{equation*}
  r_i = y_i - M_{1,1} \left(X^i\right)
\end{equation*}
and subdivided into two even clusters over the median residual in the sorted list. At the second step, new expectation models are built for each cluster, resulting in $M_{2,1}$ and $M_{2,2}$. Then, within each cluster, the records are again re-sorted according to the residuals. Each cluster is then subdivided into two clusters each over the median residual, resulting in four clusters in total, for which new expectation models $M_{3,1}$, $M_{3,2}$, $M_{3,3}$, and $M_{3,4}$ are built. The process is continued in the similar way. 

The average error for each cluster should decline in the subdividing process. Declining errors indicate that the selected model type is good enough for the entire dataset to be represented by a collection of models --- the larger is the number of models in the collection, the better is the fit to all data points. Non-declining errors, in turn, mean that the dataset cannot be represented by a collection of models of the selected type. The latter may be due to model inadequacy.

When models become sufficiently accurate, the process stops. The ensemble of models obtained at the last subdividing step can now be used for approximating the distribution of the output for each individual record --- the outputs of the ensemble can be handled as a sample from the probability distribution of the real output. 

The proposed approach falls into the boosting category, since a series of models is built sequentially were the next group of models depends on the previous. It has similarities to some known methods, but in the form proposed here, it is novel to the best knowledge of the authors. A possibly close concept is called \emph{anti-clustering} \cite{Valev1983,Spath1986,Papenberg2021}, where the data is clustered such that the clusters have maximum similarity, while having maximum diversity of the elements within them.

\subsection{Schematic illustration of DDR} 
\label{subsec:DDRExplained}

\begin{figure}
	\begin{center}
		\includegraphics{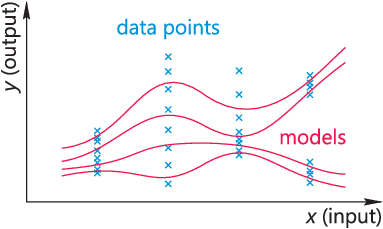}
	\end{center}
	\caption{A schematic illustration of the concept of the proposed algorithm. The ensemble of models is built for obtaining the probabilistic properties of the output.}
	\label{fig:example}
\end{figure}

The multi-modality detection by the proposed algorithm is illustrated schematically in figure \ref{fig:example}. The columns of different output values are multiple outputs corresponding to a single input. The red lines are the expectation models in the ensemble. The line `densities' within each column approximately match the `densities' of the output values, which follows from the construction of the ensemble of the models --- the first model (not illustrated) will use all data points and will pass close to mean values for each column, the data points will be subdivided into two clusters (e.g. `above' and `below' the model), and the subsequent models will be built only using the data points of the corresponding cluster. 


It should be noted that expectation models with the output continuously-dependent on the input require the probability distribution of the output of the modelled system to be continuously-dependent on the input. The figure shows four lines only to illustrate the concept; the subdividing steps in the DDR algorithm continue until the errors become sufficiently small. 

\subsection{Relation to k-nearest neighbours (kNN)} 
\label{subsec:relation2kNN}

In general, kNN is considered as a competitive method for various types of probabilistic modelling \cite{Fathabadi2022}. For example, it can be applied to the scenario illustrated in figure \ref{fig:example}. If a given input matches any of the inputs associated with the columns of data points, application of the kNN algorithm with a proper choice of the number of nearest neighbours will directly give the collection of the points from the column. The latter is close to the set of outputs of the models comprising the DDR ensemble. However, the figure shows an ideal scenario only for explanation of the concept. In reality, the points are spread along the input axes and there are gaps in the data. Furthermore, the applicability of kNN cannot always be determined from the data --- even for a deterministic system, the neighbouring points can be collected and variances can be estimated, which may lead to a wrong conclusion.

Thus, the similarity between DDR and kNN consists in returning a pre-defined number of outputs, which in the ideal case (e.g. figure \ref{fig:example}), are similar or identical. These outputs represent the scatter of points corresponding to one input point, given that the considered system is stochastic. However, the DDR's outputs are the results of the models, while kNN's outputs are the values directly picked from the dataset.

DDR frees researchers from the difficult choices related to kNN, such as the number of neighbours to pick, the weights to be assigned for the points that are slightly off, how to cover the gaps in the data, the applicability of kNN in the first place (whether the object is deterministic or stochastic). The most important advantage of DRR is that DDR does not require re-sorting of entire dataset for each prediction. The numerical comparison of DDR and kNN is provided below.

\subsection{Prediction of probability density} 
\label{subsec:density}

When the residuals decline sharply during the DDR steps, the number of obtained clusters can be small, e.g. $8$ or $16$. The modelling goal, however, is to predict the probability distribution, which requires a relatively large ensemble. Some approximation methods, such as kernel density estimation (KDE) \cite{Parzen1962}, can be applied, but DDR allows significant enlargement of the size of the ensemble, which can be used to build empirical cumulative distribution functions (ECDFs).

When clustering process is completed, any large enough group of adjacent records in the sorted list can be used for building an additional model for the ensemble. This is possible because each cluster contains records that already have been sorted at the preceding step of the algorithm. Building additional models for the existing ensemble can be one option.

The second option is constructing a different (additional) ensemble of models using a \emph{sliding window} technique, which is as follows. The finally-sorted dataset is considered. Exactly $k$ sequential records are selected, starting from the first record, and are used to build the first model for the additional ensemble. Then, $k$ sequential records starting from record $\left(s+1\right)$ are used to build the second model for the additional ensemble. This is repeated in the similar way. Such process can be interpreted as having a window of a selected size that is moved along the finally-sorted dataset, and the records that fall into this window are used for training new models for the additional ensemble. When $s<k$, the window is moved with overlapping. The users must make a choice regarding size $k$ and shift $s$, but using some knowledge of the residual errors during the clustering process, this is not a hard choice.

\section{Kolmogorov-Arnold model}
\label{sec:PKAN}

The Kolmogorov-Arnold model (or network, KAN) is given by the following expression:
\begin{equation}
\hat{y}_i = \sum_{k=1}^{n} \varPhi^k \left( \sum_{j=1}^{m} f^{kj} \left(X^i_j\right) \right) ,
\label{eq:KAR}
\end{equation}
where scalar $\hat{y}_i$ is the calculated model output of the $i$-th record, scalars $X^i_j$ denote the $j$-th component of input vector $X^i$, functions $f^{kj}$ and $\varPhi^k$ constitute the model. In the original work \cite{Arnold1957,Kolmogorov1956}, it has been shown that the model with $n=2m+1$ can represent any continuous multivariate function\footnote{More recently the restrictions on continuity have been somewhat relaxed, e.g. see discussion in \cite{Ismailov2008}.}. Functions $f^{kj}$ and $\varPhi^k$, which are referred to as the `inner' and the `outer' functions, respectively, are further decomposed into basis functions and parameters. There are multiple model training methods and choices for the basis functions. It is assumed that the reader is already familiar with this model; an overview of the model and its training methods can be found in the recent publication by the authors \cite{Poluektov2023}; practical examples of the model application can be found on the authors' website\footnote{http://openkan.org}.

\subsection{Shallow probabilistic model}
\label{sec:shallowPKAN}

For relatively large datasets, entire KAN can be used as the expectation model within the DDR algorithm. However, for relatively small datasets that might not be possible --- subdivision of the dataset into clusters may lead to such cluster size that the number of records within a cluster is insufficient for model training without overparametrisation. The latter can be addressed using the idea as follows.

Model \eqref{eq:KAR} can be rewritten as
\begin{align}
  &\hat{y}_i = \sum_{k=1}^{n} \varPhi^k \left( \theta_k^i \right) , \label{eq:GAM} \\
  &\theta_k^i = \sum_{j=1}^{m} f^{kj} \left(X^i_j\right) ,
\end{align}
where intermediate variable $\theta^i \in \mathbb{R}^n$ has been introduced, with index $i$ indicating the record number. The combination of DDR and KAN consists in fitting a single expectation model (KAN) for entire dataset first. Then, the model is used to calculate the values of intermediate variables $\theta^i$. These values are then assembled into a new dataset $\left( \theta^i, y_i \right)$, $i \in \left\lbrace 1, \ldots, N \right\rbrace$. Finally, the DDR algorithm is applied to the new dataset with the expectation model having the form given by equation \eqref{eq:GAM}.

Such approach allows reducing significantly the number of estimated parameters within a single expectation model in the ensemble, also leading to a faster training (compared to using KANs as expectation models). Model \eqref{eq:GAM} is called a \emph{generalised additive model} (GAM) --- it can be viewed as a constituent part of KAN, and for further information on the relation between these models, the reader is referred to \cite{Poluektov2023}.

\section{Elementary example --- application of DDR}
\label{sec:comparison}

The aim of the first test is to demonstrate the capabilities of DDR using a simple expectation model (without the combination with KAN). All computational tests of the present paper use synthetic data, as the reference (true) distributions are required for the validation; in the case of experimentally-obtained datasets of stochastic systems, true distributions are unknown.

\subsection{Data and expectation model}
\label{sec:dataset}

Synthetic data is generated using the Monte Carlo (MC) algorithm. The data corresponds to a stochastic system, the output of which is the sum of outcomes of $q$ ten-sided dice rolls. There are three inputs: $q_1$, $q_2$, and $p$, where $1 \leq q_{1,2} \leq 10$. The number of dice is then chosen as $q=q_1$ with probability $p$, and $q=q_2$ with probability $\left(1-p\right)$. The output of such system is a discrete random variable, the distribution of which is input-dependent and is also bi-modal for a large range of inputs. The training and the validation datasets consist of $1000$ and $100$ records, respectively; the inputs are drawn independently from a uniform distribution; the validation dataset is generated independently from the training dataset. 

Seven steps of the DDR algorithm are done, resulting in $2^{7-1} = 64$ expectation models in the DDR ensemble. Afterwards, the sliding window technique is applied with the window size of $20$ records, which is moved along the sorted records without overlapping, providing the final ensemble of $50$ models. The sliding window method is not necessary for this particular dataset, but its usage is a part of the test.

For this example, the most simple expectation model is chosen --- a multilinear (trilinear) model:
\begin{equation}
\hat{y}^i = c_0 + c_1 X^i_1 + c_2 X^i_2 + c_3 X^i_3 + c_4 X^i_1 X^i_2 + c_5 X^i_1 X^i_3 + c_6 X^i_2 X^i_3 + c_7 X^i_1 X^i_2 X^i_3 ,
\label{eq:bilinear}
\end{equation}
where $\hat{y}^i$ is the estimated output of the $i$-th record, $X^i_j$ denote the $j$-th component of input vector, $c_j$ are the model parameters. 

\subsection{Accuracy metrics}
\label{sec:metrics}

To assess the performance of the approach, the modelling results are compared to the MC simulations that can be considered to represent true distributions. 

The first comparison step considers means and standard deviations obtained from the DDR ensemble (modelling) and from the MC simulations (reference solution). The normalised root mean square error (RMSE) measure is used:
\begin{equation}
  E_\mathrm{RMSE} = \frac{1}{\operatorname{max}\left(Z_\mathrm{MC}\right) - \operatorname{min}\left(Z_\mathrm{MC}\right)}\sqrt{\frac{1}{N}\sum_i\left(Z_\mathrm{DDR}^i - Z_\mathrm{MC}^i\right)^2} ,
\label{eq:RMSE}
\end{equation}
where $Z$ stands for either mean or standard deviation, subscript `$\mathrm{DDR}$' denotes the value obtained from the DDR ensemble, subscript `$\mathrm{MC}$' denotes the value obtained from the MC sampling, $N$ is the number of records used for the comparison, and the sum is over record number $i$.

Furthermore, goodness-of-fit comparisons are also performed. As the output is a discrete random variable, a test similar to the Cram\'{e}r-von-Mises test is used. The statistic is the relative distance between the \emph{median trees} constructed as follows. Two samples (modelling and reference solution) are compared. Each sample is subdivided by the median into two clusters; this median is the first entry in the list; another two medians for each of the obtained clusters are added to the list; the process continues in the same way until the predefined number of the clusters' medians is obtained. The obtained medians are sorted, resulting in a vector that can be called the \emph{median tree}. The vectors corresponding to the two samples ($U$ and $V$) are compared using the relative distance:
\begin{equation}
  S = 2 \sqrt{2} \frac{\lVert U - V \rVert}{\lVert U \rVert + \lVert V \rVert} .
\end{equation}
It is easy to show that if two samples correspond to the same continuous random variable with probability density function $f$, then the numerator of $S$ is related to statistic $T$ from the Cram\'{e}r-von-Mises test \cite{Anderson1962}. In particular, statistic 
\begin{equation}
  T_* = \frac{a b}{c\left(a+b\right)} \sum_i \left( \left( U_i - V_i \right) f\left( \frac{U_i + V_i}{2} \right) \right)^2 ,
\end{equation}
where $a$ and $b$ are the sizes of two samples, $c$ is the number of medians (elements in $U$ or $V$), has the same limiting distribution as $T$ when $a \to \infty$, $b \to \infty$, and $a/b \to \lambda$, where $\lambda$ is finite. Statistic $S$ simply omits $f$ and is scaled differently to $T_*$. Theoretical limits for $S$ are $0 \leq S \leq 2 \sqrt{2}$, but the expected value of $S$ for two random $U$ and $V$ (assuming the uniform distribution with the same limits) is $1$; therefore, this value is a convenient measure with conventional range $[0,1]$. 

The standard approach for most similar testing procedures is to compare the statistic computed for two samples to tabulated values. Since this is a new test, such tables are not available, but can be created \emph{ad hoc}. The MC sample can be as large as necessary and represents the population, against which the model sample is tested. Random sub-samples from the MC population are taken, statistic $S$ is calculated for each taken sub-sample and the MC population, the values are sorted and used as an ECDF for the validation. For the examples of this paper, the number of sub-samples is taken to be $100$, and when the statistic for the model sample is below the maximum value, the test is considered to be passed at $1\%$ significance level. 

\subsection{Test results}
\label{sec:results}

The results are shown in table \ref{DDR}. Each column is obtained by independent execution of the programme, which includes the data generation, the training, and the validation (against MC) as described above. For this example, the normalised RMSEs for the mean and for the standard deviation obtained from the DDR ensemble are approximately $4\%$ and $14\%$, respectively. The samples given by the DDR ensemble pass approximately $85$ out of $100$ goodness-of-fit tests.

The last row of table \ref{DDR} shows the application of kNN to the example above. The number of neighbours is the same as the number of models in the DDR ensemble. It can be seen that kNN samples pass fewer goodness-of-fit tests, approximately $73$ out of $100$.

\begin{table}
\begin{center}
	\begin{tabular}{| c | c c c c c c c c |} 
		\hline
		Sequential number & 1 & 2 & 3 & 4 & 5 & 6 & 7 & 8 \\ 
		\hline
		RMSE for mean & 0.04 & 0.03 & 0.05 & 0.03 & 0.04 & 0.05 & 0.04 & 0.03 \\ 
		\hline
		RMSE for standard deviation & 0.15 & 0.14 & 0.14 & 0.15 & 0.14 & 0.14 & 0.14 & 0.15 \\ 
		\hline
		Passed goodness-of-fit tests, DDR & 79 & 91 & 80 & 92 & 86 & 77 & 89 & 87 \\ 
		\hline
		Passed goodness-of-fit tests, kNN & 62 & 81 & 70 & 62 & 77 & 73 & 83 & 73 \\ 
		\hline
	\end{tabular}
	\caption{Test results for the dice example.}
	\label{DDR}
\end{center}
\end{table} 

\begin{figure}
	\begin{center}
		\includegraphics{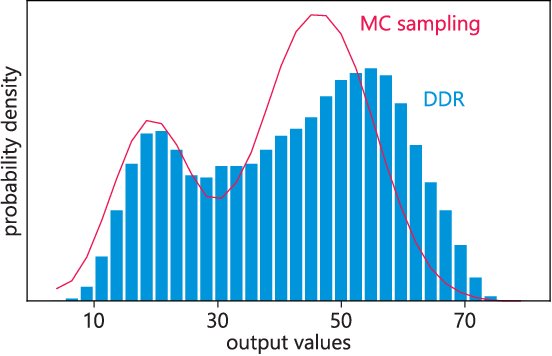}
	\end{center}
	\caption{Probability density of the output for the dice example: the reference solution obtained using the Monte Carlo (MC) sampling (solid line) and the estimation made using the divisive data re-sorting (DDR) algorithm (bar chart).}
	\label{fig:D349}
\end{figure}

Estimation of the probability density using KDE applied to the DDR ensemble is shown in figure \ref{fig:D349} for $p=0.3$, $q_1=4$, $q_2=9$, where it can be seen that the bi-modal character of the distribution is captured. The method consists in replacing a single value in a sample by a smooth distribution curve (e.g. Gaussian) and obtaining the distribution as a result of superposition. The expected smoothness of the distribution is controlled by a numerical parameter.


Table \ref{Blocks} shows a sharp decline of the relative error in the subdividing process of the DDR algorithm computed during training on the training dataset itself, confirming the aleatoric nature of the uncertainty. The source code is available online\footnote{https://github.com/andrewpolar/vdice\_bilinear}.

\begin{table}
\begin{center}
	\begin{tabular}{| c | c c c c c c c |} 
		\hline
		Number of clusters & 1 & 2 & 4 & 8 & 16 & 32 & 64 \\ 
		\hline
		RMSE for output & 0.153 & 0.107 & 0.058 & 0.032 & 0.018 & 0.009 & 0.004 \\ 
		\hline
	\end{tabular}
	\caption{Decline of the relative error in the DDR algorithm with increase of the number of clusters.}
	\label{Blocks}
\end{center}
\end{table}

\subsection{Comparison to Bayesian neural networks (BNNs)}
\label{sec:BNN}

The multi-modal input-dependent probability distributions of the outputs of a stochastic system can also be estimated using BNNs \cite{JerfelG}. A well-tested and popular code Keras\footnote{https://keras.io/examples/keras\_recipes/bayesian\_neural\_networks} is taken for the comparison. Since the reference input-dependent distributions should be known (i.e. needed for the validation), in the provided example\footnote{https://archive.ics.uci.edu/ml/datasets/wine+quality}, the experimentally-obtained dataset has been replaced by the described-above synthetic dataset. The code has been used almost as is, with only one necessary change --- to support multi-modality, the posterior (output) distribution type has been changed from a single normal distribution to a mixture of two normal distributions, e.g.
\begin{equation}
  f\left(x\right) = \sum_{j} p_j f_\mathcal{N} \left( x; \mu_j, \sigma^2_j \right) ,
\label{eq:mixture}
\end{equation}
where weights $p_j$, expectations $\mu_j$, and standard deviations $\sigma_j$ are estimated from the data. The latter is a typical way of incorporating multi-modality in probabilistic neural networks \cite{Mohebali2020}.



The results are shown in table \ref{BNN_Dice}. It can be seen that it is less accurate than DDR for this particular dataset. The reason is a relatively small dataset size, but it is a part of the test. BNN also captures multi-modality, which can be shown by making the dataset less challenging --- by increasing the size, excluding uni-modal records, or narrowing the variation ranges for the inputs, e.g. $3 \leq q_{1,2} \leq 6$. The source code is available online\footnote{https://github.com/andrewpolar/vdice-python}.

\begin{table}
\begin{center}
	\begin{tabular}{|c c c c c c c c c|} 
		\hline
		Sequential number & 1 & 2 & 3 & 4 & 5 & 6 & 7 & 8\\ 
		\hline
		RMSE for mean & 0.19 & 0.17 & 0.19 & 0.14 & 0.15 & 0.18 & 0.15 & 0.17\\ 
		\hline
		RMSE for standard deviation & 0.20 & 0.21 & 0.30 & 0.43 & 0.43 & 0.26 & 0.32 & 0.23\\ 
		\hline
		Passed goodness-of-fit tests & 19 & 23 & 18 & 16 & 13 & 25 & 23 & 23\\ 
		\hline
	\end{tabular}
	\caption{Test results for the dice example, modelled using BNN.}
	\label{BNN_Dice}
\end{center}
\end{table}

\section{Probabilistic KAN test}
\label{sec:second}

The toy dataset in the previous example allowed using an elementary polynomial expectation model. This example introduces a more complex and closer to reality dataset that cannot be modelled by simple models, requiring a complex network-type model with multiple layers. The data is computed using the following formula:
\begin{equation}
\begin{split}
  &y = \frac{2 + 2 X_3^*}{3\pi} \left( \operatorname{arctan}\left( 20\left( X_1^* - \frac{1}{2} + \frac{X_2^*}{6}\right)\exp\left(X_5^*\right) \right) + \frac{\pi}{2} \right) + \\ 
  &\vphantom{y} + \frac{2 + 2 X_4^*}{3\pi} \left( \operatorname{arctan}\left( 20\left( X_1^* - \frac{1}{2} - \frac{X_2^*}{6}\right)\exp\left(X_5^*\right) \right) + \frac{\pi}{2} \right) , \\
  &X_j^* = X_j + \delta \left(C_j - 0.5\right) , \quad j \in \left\lbrace 1, \ldots, 5 \right\rbrace ,
\end{split}
\label{eq:formula2}
\end{equation}
where $y$ is the system output, $X_j$ are the system inputs, $C_j \sim \operatorname{unif}\left(0,1\right)$ are uniformly distributed random variables (noise), multiplier $\delta = 0.4$ is the aleatoric uncertainty level (the system becomes deterministic for $\delta = 0$). For data generation, inputs $X_j \sim \operatorname{unif}\left(0,1\right)$ are taken.

\begin{figure}
  \begin{center}
    \includegraphics{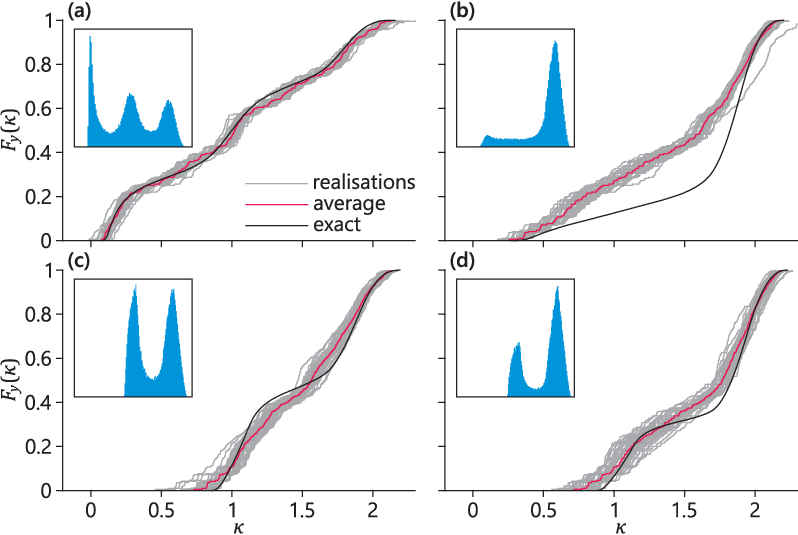}
  \end{center}
  \caption{The empirical cumulative distribution functions (ECDFs) of the output of the considered stochastic system obtained using the Monte Carlo (MC) sampling (black) and using the realisations of the divisive data re-sorting (DDR) algorithm (grey). Averages over the realisations are shown in red. The corresponding probability density functions (PDFs) obtained using the MC sampling are shown in the insets. Subfigures (a)-(d) correspond to inputs $X^1$-$X^4$.}
  \label{fig:atanGeom}
\end{figure}

The probability density of the output significantly changes depending on the inputs. In figure \ref{fig:atanGeom}, in the insets (blue figures), four examples of probability densities of $y$ are shown for four different inputs:
\begin{align*}
  &X^1 = \begin{bmatrix} 0.5 & 0.5 & 0.5 & 0.5 & 0.5 \end{bmatrix} , \\
  &X^2 = \begin{bmatrix} 0.65 & 0 & 0.5 & 0.5 & 0.5 \end{bmatrix} , \\
  &X^3 = \begin{bmatrix} 0.68 & 1 & 0.5 & 0.5 & 0.5 \end{bmatrix} , \\
  &X^4 = \begin{bmatrix} 0.74 & 1 & 0.5 & 0.5 & 1 \end{bmatrix} ,
\end{align*}
corresponding to subfigures (a)-(d), respectively. The probability density functions (PDFs) are built using the MC sampling of $10^5$ points.

To benchmark the DDR procedure, a total of $40$ runs of the programme have been performed. During each run, a dataset of $10^6$ records has been generated, the ensemble of models has been constructed using the DDR algorithm, the ECDFs for the given above four points have been calculated using the sliding window technique. The number of the outer functions in the Kolmogorov-Arnold model has been selected to be $11$; the inner and the outer functions have been taken to be piecewise linear with $5$ and $7$ equidistant nodes, respectively. 

In figure \ref{fig:atanGeom}, the given above four inputs are considered, the ECDFs of the output built using the MC sampling are shown in black colour and the ECDFs obtained using the DDR procedure (with the sliding window technique) are shown in grey colour\footnote{The output of the system, denoted as $y$ in equation \eqref{eq:formula2}, is a random variable. Hence, the ECDF of $y$ is denoted as $F_y$, and its argument is denoted as $\kappa$, which stands for a value that $y$ may take.}. Furthermore, an average ECDF across all realisations (i.e. average of grey curves) is shown in red colour. For the purpose of the discussion, the MC sampling ECDFs are referred to as the exact. It can be seen that the ensemble of models reproduces the major features of the exact CDFs qualitatively and predicts well the range of the output quantitatively. The subfigures show different representative scenarios: for input $X^1$, the ensemble ECDF is very close to the exact CDF; for input $X^2$, the ensemble ECDF is somewhat far from the exact CDF quantitatively, although reproduces qualitatively the change of the slope; for inputs $X^3$ and $X^4$, the ensemble ECDF is rather close to the exact CDF, although somewhat smoothed. 

\begin{figure}
  \begin{center}
    \includegraphics{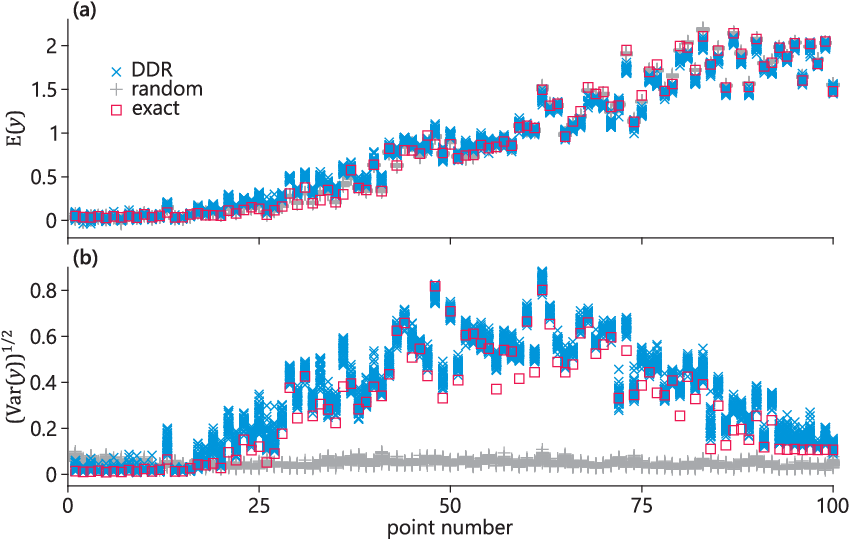}
  \end{center}
  \caption{The mean and the standard deviation for $100$ input points. The points are sorted based on their $X_1$ value. The values are obtained using the MC sampling (red squares), using the DDR ensembles (blue crosses), and using the ensembles of models trained on randomly selected disjoint sets of records (grey pluses).}
  \label{fig:atanMV}
\end{figure}

The DDR algorithm builds the ensemble of models using the specifically created set of data clusters. Therefore, it is crucial to emphasise its advantage over an ensemble built using clusters containing random records. To show this, a set of $100$ input points have been randomly generated. For each input point, the true mean and the true standard deviation have been calculated using the MC sampling. Next, for each input point, the mean and the standard deviation have been calculated using the DDR ensembles and the random ensembles\footnote{The entire dataset is shuffled randomly, split into the same number of clusters as the DDR ensemble contains, and the ensemble of models is created  such that each model is trained on its cluster.}. In figure \ref{fig:atanMV}, the results are compared and it can be seen that the random ensembles can model the means well, which is a known fact, but cannot model the standard deviations at all. Meanwhile, the DDR ensembles predict the standard deviations both qualitatively and quantitatively.

\subsection{Shallow probabilistic KAN}
\label{sec:shallow}

Following section \ref{sec:shallowPKAN}, a shallow model is constructed for the example above. The number of records is decreased to $10^4$. The structure of the Kolmogorov-Arnold model has been changed to $6$ and $12$ equidistant nodes for the inner and the outer functions, respectively (as above, piecewise linear functions are used). Five steps of the DDR algorithm are done, resulting in $2^{5-1}=16$ clusters. For the training of the DDR ensemble models, the number of equidistant nodes for the outer functions is decreased to $7$. The sliding window of $500$ records with the shift of $300$ records is used, providing the final ensemble of $32$ models. Thus, each individual model of the final ensemble containing $77$ parameters (since there are $11$ outer functions with $7$ nodes per function) is trained on $500$ records (the sliding window size). 
 
The same accuracy metrics as in the dice example are used, and the results are shown in table \ref{PKAN}. The errors for the mean and for the standard deviation are even lower than for the dice example; the number of passed goodness-of-fit tests is lower, but accounting for the complexity of the input-dependent distributions (shown in the insets of figure \ref{fig:atanGeom}), the authors consider this result to be acceptable. The source code is available online\footnote{https://github.com/andrewpolar/pkan}.

One common problem for probabilistic models is a relatively long training time (e.g. several minutes even for relatively small datasets). Construction of the shallow probabilistic KAN using DDR in this example took approximately $0.25$ seconds on Intel(R) Core(TM) i7-8550U 1.80 GHz CPU, which can be considered to be a relatively good result. This can be crucial for cases focusing on unsupervised learning using large datasets typically requiring hours and days for training. 


\begin{table}
\begin{center}
	\begin{tabular}{| c | c c c c c c c c |} 
		\hline
		Sequential number & 1 & 2 & 3 & 4 & 5 & 6 & 7 & 8 \\ 
		\hline
		RMSE for mean & 0.04 & 0.03 & 0.04 & 0.03 & 0.04 & 0.03 & 0.04 & 0.03 \\ 
		\hline
		RMSE for standard deviation & 0.10 & 0.13 & 0.12 & 0.16 & 0.10 & 0.14 & 0.12 & 0.13 \\ 
		\hline
		Passed goodness-of-fit tests & 58 & 58 & 62 & 60 & 64 & 63 & 63 & 64 \\ 
		\hline
	\end{tabular}
	\caption{Test results for shallow probabilistic KAN.}
	\label{PKAN}
\end{center}
\end{table}

\section{Conclusions}
\label{sec:conclusion}

The present paper has introduced a new approach to modelling of systems with aleatoric uncertainty --- the divisive data re-sorting (DDR) method. It consists in recursive steps of training of an ensemble of models, each on an individual cluster of data records, sorting of the records, and increasing of the ensemble size. The method can be combined with any expectation model, but the model should have adequate descriptive capabilities for the considered data. The suggested method can capture the multi-modality of the input-dependent probability distribution of the system's output, as numerical experiments have shown. The method can also be used when users do not need a distribution and are satisfied with first two statistical moments for the computed outputs. The method is related to kNN but is free from many of its hard choices, as well as from sorting of the entire dataset for each prediction.

The DDR method has been then combined together with the Kolmogorov-Arnold model (or network, KAN) resulting in the shallow probabilistic KAN. It has been found to be an efficient combination of a regression model with high descriptive capabilities and a simple model (the generalised additive model, or GAM) as a component of the DDR ensemble for producing the probabilistic output. Its relatively quick training may be crucial for some applications. 


\bibliographystyle{unsrt}
\bibliography{refs}

\end{document}